\documentclass[conference]{IEEEtran}
\IEEEoverridecommandlockouts

\usepackage{cite}
\usepackage{amsmath,amssymb,amsfonts}
\usepackage{algorithmic}
\usepackage{graphicx}
\usepackage{textcomp}
\usepackage{xcolor}
\usepackage{float}
\usepackage{hhline}
\usepackage{booktabs}
\usepackage{tabularx}
\usepackage{lineno}
\usepackage{hyperref}
\usepackage{amsmath}

\def\BibTeX{{\rm B\kern-.05em{\sc i\kern-.025em b}\kern-.08em
    T\kern-.1667em\lower.7ex\hbox{E}\kern-.125emX}}
\makeatletter
\newcommand{\linebreakand}{%
\end{@IEEEauthorhalign}
\hfill\mbox{}\par
\mbox{}\hfill\begin{@IEEEauthorhalign}
}
\makeatother
\begin{document}

\title{Perturbation on Feature Coalition:\\ Towards Interpretable Deep Neural Networks}

\author{
	\IEEEauthorblockN{1\textsuperscript{st} Xuran Hu}
	\IEEEauthorblockA{\textit{School of Electronic Engineering} \\
		\textit{Xidian University}\\
		Xi'an, China \\
		XuRanHu@stu.xidian.edu.cn}
	\and
	\IEEEauthorblockN{2\textsuperscript{nd} Mingzhe Zhu}
	\IEEEauthorblockA{\textit{School of Electronic Engineering} \\
		\textit{Xidian University}\\
		Xi'an, China \\
		zhumz@mail.xidian.edu.cn}
	\and
	\IEEEauthorblockN{3\textsuperscript{rd} Zhenpeng Feng}
	\IEEEauthorblockA{\textit{School of Electronic Engineering} \\
		\textit{Xidian University}\\
		Xi'an, China \\
		zpfeng\_1@stu.xidian.edu.cn}
	\linebreakand
	\IEEEauthorblockN{4\textsuperscript{th} Milo\v{s} Dakovi\'c}
	\IEEEauthorblockA{\textit{Faculty of Electrical Engineering} \\
		\textit{University of Montenegro}\\
		Podgorica, Montenegro \\
		milos@ucg.ac.me}
	\and
	\IEEEauthorblockN{5\textsuperscript{th} Ljubi\v{s}a Stankovi\'c}
	\IEEEauthorblockA{\textit{Faculty of Electrical Engineering} \\
		\textit{University of Montenegro}\\
		Podgorica, Montenegro \\
		ljubisa@ucg.ac.me}
}


\maketitle

\begin{abstract}
The inherent "black box" nature of deep neural networks (DNNs) compromises their transparency and reliability. Recently, explainable AI (XAI) has garnered increasing attention from researchers. Several perturbation-based interpretations have emerged. However, these methods often fail to adequately consider feature dependencies. To solve this problem, we introduce a perturbation-based interpretation guided by feature coalitions, which leverages deep information of network to extract correlated features. Then, we proposed a carefully-designed consistency loss to guide network interpretation. Both quantitative and qualitative experiments are conducted to validate the effectiveness of our proposed method. Code is available at \href{https://github.com/Teriri1999/Perturebation-on-Feature-Coalition}{github.com/Teriri1999/Perturebation-on-Feature-Coalition}.
\end{abstract}


\section{Introduction}

Deep neural networks (DNNs) have achieved widespread application across various fields. However, their black-box nature hinders our understanding of their decision-making processes, which limits their usage in sensitive areas such as healthcare, finance, and autonomous driving. Moreover, some studies suggest that DNNs may exhibit "Clever Hans" phenomenon \cite{lapuschkin2019unmasking}, further undermining their reliability.

To enhance the transparency of DNNs, numerous network interpretation \cite{hu2024manifold} have been proposed. Perturbation-based methods \cite{zhu2024unveiling} offer an intuitive way to assess the impact of input features on network decisions. Initially, methods like Occlusion \cite{zeiler2014visualizing} involved traversing and masking image pixels to determine feature importance based on confidence drop in networks' forward propagation. However, these approaches often overlook feature interdependencies, potentially leading to incorrect interpretations. Subsequently, more advanced methods \cite{ivanovs2021perturbation, petsiukrise} have incorporated optimization techniques by masking features through random sampling and optimizing the mask to minimize confidence drop. While this approach has shown some success, it still faces challenges in ensuring mask sparsity during optimization, potentially leading to local optima and neglecting feature interdependencies.

To address these issues, this paper proposes a perturbation-based interpretation on feature coalitions, which involves two main steps: (1) extracting deep information of network to identify correlated features and (2) using these correlated features to guide perturbation interpretation. Specifically, for the network to be explained, we retain only the convolutional layers, activation layers, and normalization layers to build an aligned feature extraction network. We train and optimize this network to extract correlated features from samples. Subsequently, we propose a carefully designed loss function to ensure consistency of explanation within correlated feature coalition, resulting in effective network interpretations.

The main contributions of this paper are as follows:

\begin{enumerate}
	\item We provide a noval features extraction approach from network's perspective, rather than relying on manually defined guidance for interpretations in previous work \cite{bitton2022latent}.
	\item We propose a well-designed regional consistency loss to guide network interpretation, ensuring the validity and robustness of model interpretation.
\end{enumerate}

\section{Methodology}

\begin{figure*}[!t]
	\centering
	\includegraphics[width=0.90\textwidth]{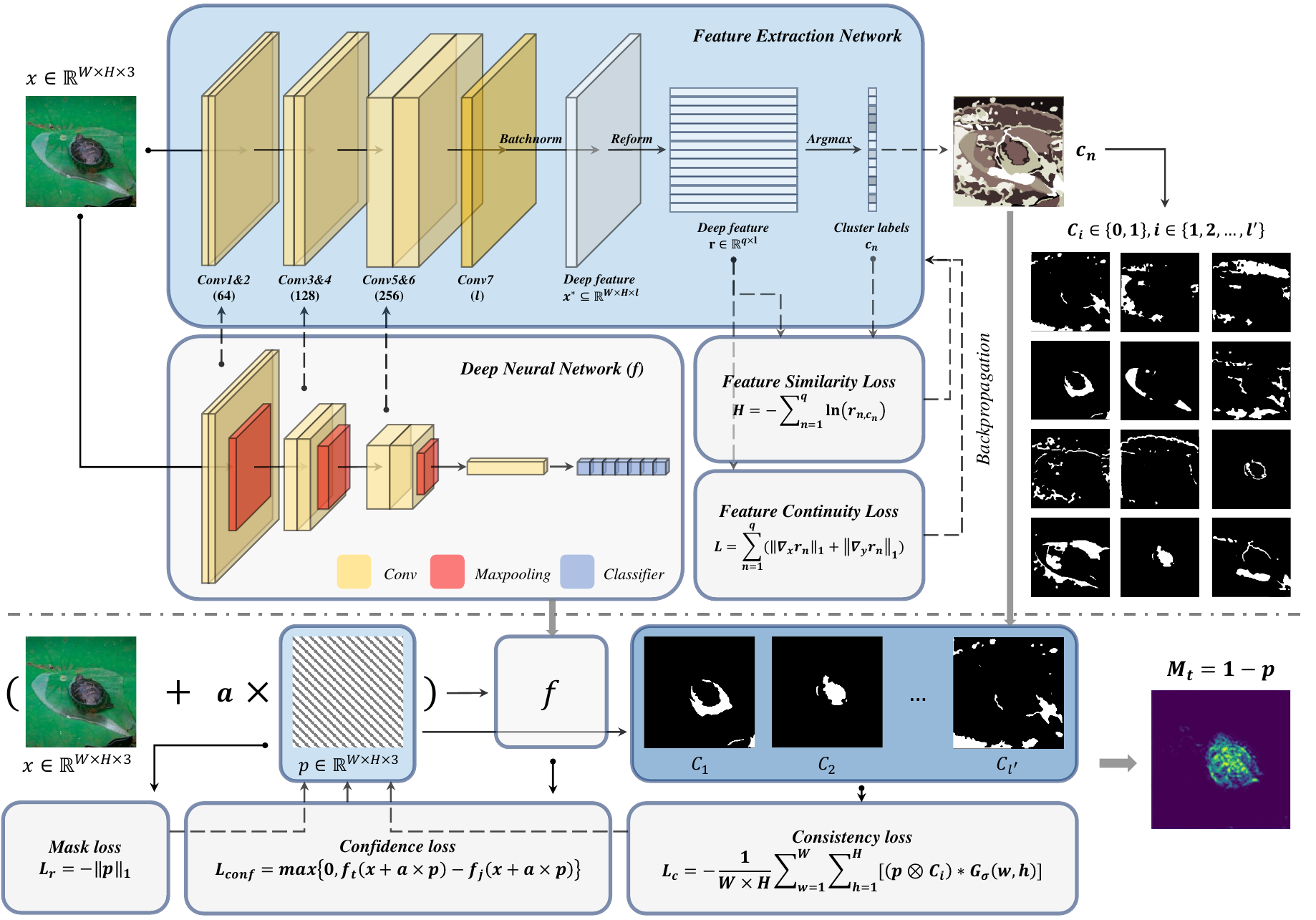}
	\caption{Implementation of the proposed method. Upper part: extraction of correlated feature coalitions; Lower part: coalition-guided perturbation interpretation.}
	\label{fig_1}
\end{figure*}

This section introduces the implementation detail of our proposed method, which involves two main steps: (1) extraction of correlated features and (2) perturbation on feature coalitions. Figure \ref{fig_1} illustrates the workflow of our method.

\subsection{Correlated Feature Extraction}

In the first step, we aim to identify correlated features within the sample. It's important to note that this correlated feature extraction is completely unsupervised, thereby avoiding introducing any subjective bias. While some previous approaches defined correlated features manually \cite{bitton2022latent}, our approach seeks to uncover feature correlations from the network's perspective, leading to more effective guidance for network interpretation.

Specifically, we introduce two criteria: (1) similar features should belong to the same category, and (2) spatially contiguous features should belong to the same category. These two principles will be used later to design loss function.

Let $f$ be a pretrained network and $x \subseteq \mathbb{R}^{W \times H \times 3}$ be a sample to interpret. Our objective is to identify the correlated feature coalitions within $x$. Inspired by previous research \cite{kim2020unsupervised}, we reconfigure network $f$, retaining only convolutional layers, activation layers, and normalization layers. Additionally, we add a feature extraction layer at the end of network, using a $1 \times 1$ convolution layer to adjust channel number to $l$, and then followed it by a batch normalization layer. The output dimension is $x^* \subseteq \mathbb{R}^{W \times H \times l}$. We then reformat this output by merging the first two dimensions to obtain $r \in \mathbb{R}^{q \times l}$, where $q=W \times H$. Intuitively, each feature $r_n$ (where $n \in\{1,2, \ldots, q\}$) is represented as a feature vector $\mathbb{R}^{1 \times l}$. From another perspective, $l$ also represents the number of clusters. Here, we simply determine which class each pixel belongs to by identifying the maximum value index in each feature vector, which can be expressed as:

\begin{equation}
	c_n=\left\{i \mid r_{n, i}>r_{n, j}, \forall j\right\},
\end{equation} where $r_{n,i}$ represents the $i$-th element of the feature vector $r_n$.

Based on the two aforementioned criteria, we define two loss functions.

\noindent \textbf{Feature Similarity Loss: }We use cross-entropy loss between cluster category $c_n$ and feature $r_n$ to ensure that similar features are assigned to the same category. The mathematical expression can be:

\begin{equation}
	H\left(r_n, c_n\right)=-\sum_{n=1}^q \ln \left(r_{n, c_n}\right).
\end{equation}

\noindent \textbf{Feature Continuity Loss: }To enforce continuity, we first reshape $r_n$ to $r_n \in \mathbb{R}^{W \times H \times l}$. We then compute the gradients of deep features along horizontal and vertical directions to ensure smoothness. The mathematical expression is as follows:

\begin{equation}
	L\left(r_n\right)=\sum_{n=1}^q\left(\left\|\nabla_{\mathbf{x}} r_n\right\|_1+\left\|\nabla_{\mathbf{y}} r_n\right\|_1\right).
\end{equation}

The total loss function is a combination of two losses: $L_t=H\left(r_n, c_n\right)+\lambda L\left(r_n\right)$. We then optimize the model to obtain correlated feature coalitions. During the optimization process, number of clusters $l$ decreases adeptly, resulting in a final number of feature coalitions $l^{\prime}$ within range $l^{\prime} \in[k, l]$, where $k$ is the minimum number of feature coalitions we specify.

\subsection{Perturbation on Feature Coalition}

In the second step, we use the feature correlation information obtained before to guide the perturbation interpretation. Let $x$ be the input sample, $f(x)$ the network output, $t=\arg\max (f(x))$ the predicted class, and mask $p$. 

\noindent \textbf{Confidence loss: }Similar to other perturbation-based methods, our goal is to find the largest perturbation $p \in P$ such that the predicted class $t=\arg \max f(x+p)$ remains unchanged. This problem can be reformulated as:

\begin{equation}
	f_t(x+p)>f_j(x+p), \forall j \neq t .
\end{equation}

Equivalently, we define the following confidence loss:

\begin{equation}
	L_{\text {conf }}=\max \left\{0, f_t(x+a \times p)-f_j(x+a \times p)\right\},
\end{equation} where $a \in[0,1]$ follows a uniform distribution.

\noindent \textbf{Mask loss: }Since we aim to find the largest mask, the mask loss can be specified:

\begin{equation}
	L_r=-\|p\|_1.
\end{equation}

\noindent \textbf{Consistency loss: }Next, we incorporate feature correlation information to guide the perturbation. We define a consistency loss over each feature coalition. From Section 2.1, we have obtained $l^{\prime}$ feature coalitions $C_i \in\{0,1\}$ for $i \in\left\{1,2, \ldots, l^{\prime}\right\}$, where the regions with a value of 1 represent the feature coalitions. We compute the consistency loss over these coalitions as follows:

\vspace{-10pt}

\begin{equation}
	L_c=-\frac{1}{W \times H} \sum_{w=1}^W \sum_{h=1}^H\left[\left(p \otimes C_i\right) * G_\sigma(w, h)\right],
\end{equation} where $\otimes$ represents element-wise multiplication and $*$ represents convolution operations. $G_\sigma$ is a Gaussian kernel function:

\begin{equation}
	G_\sigma(w, h)=\frac{1}{2 \pi \sigma^2} \exp \left(-\frac{w^2+h^2}{2 \sigma^2}\right),
\end{equation} 

Here, we use $\sigma=1.0$.

Finally, we compute total loss $L=L_r+\mu \cdot L_{\text {conf }}+v \cdot L_c$ and optimize it to obtain the maximum perturbation mask $p$ and the saliency map $M_t=1-p$.

\section{Experimental Results}

\subsection{Experiment Setup}

\noindent \textbf{Dataset:} This study utilizes ImageNet-1k dataset, which is pretrained on VGG16.

\noindent\textbf{Evaluation Metrics: }A satisfactory interpretation necessitates the inclusion of most relevant pixels pertaining to objects contributing to the decision-making process. We used three metrics for effectiveness evaluation here. First, Average drop in confidence:

\begin{equation}
	AD =\frac{1}{N} \sum_{i=1}^N \frac{f\left(x_i\right)-f\left(M_t\right)}{f\left(x_i\right)} \text {, }
\end{equation} where $N$ is the total number of samples.

Second, Percentage of increase in confidence:

\begin{equation}
		PI =
		\sum_{i=1}^N\left(\frac{m\left(f\left(x_i\right)<f\left(M_t\right)\right)}{N}\right) \text {, } 
\end{equation} where $m(\cdot) \in\{0,1\}$ is a function used to judge whether a sample satisfies the condition of increasing confidence.

And Top-1 accuracy: 

\begin{linenomath}
	\begin{equation}
		T_1 =\frac{m\left(f(M_t)_{c=t}>f(M_t)_{c \neq t}\right)}{N},
	\end{equation}
\end{linenomath}.

\noindent \textbf{Implementation Details: }In this study, we implemented our method using PyTorch framework, with a pre-trained VGG16 model from PyTorch as the network to be explained. The loss coefficients were set as follows: $\lambda = 1.0$, $\mu = 100.0$, and $v = 1.0$. For feature coalition extraction, we initialized the number of clusters at 20, with a minimum cluster count set to 4. All experiments are conducted on a platform with Intel 12700f and NVIDIA RTX3080 GPU.

\subsection{Qualitative Comparison}

\begin{figure}[!t]
	\centering
	\includegraphics[width=\linewidth]{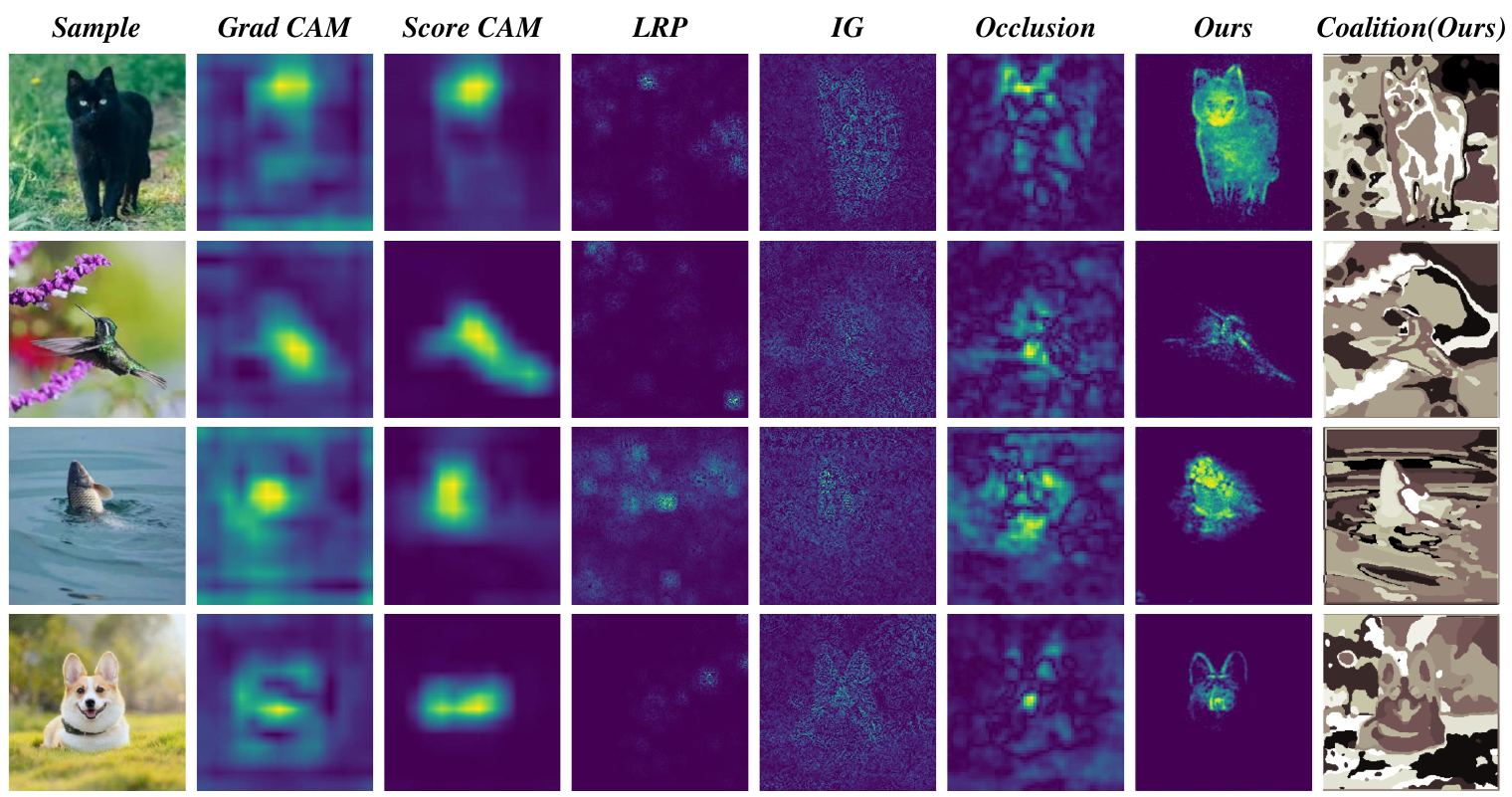}
	\caption{Comparison of interpretation methods on ImageNet-1k. Column from left to right: input sample, Grad-CAM\cite{selvaraju2017grad}, Score CAM\cite{wang2020score}, LRP\cite{bach2015pixel}, IG\cite{sundararajan2017axiomatic}, Occlusion \cite{zeiler2014visualizing}, our proposed method, and extracted feature coalition.}
	\label{fig_2}
\end{figure}

Figure \ref{fig_2} presents a quantitative comparison of different interpretations. Our proposed approach demonstrates superior performance in locating regions of network interest and recovering fine-grained features that decision-making relies on. Unlike traditional perturbation-based methods, our approach fully considers feature interdependencies, resulting in more effective visualizations.

\vspace{10pt}

\begin{table}[t] 
	\caption{Quantitatively measurement of each method.\label{tab1}}
	\scriptsize
	\centering
	\newcolumntype{C}{>{\centering\arraybackslash}X}
	\begin{tabularx}{\linewidth}{p{1.8cm}CCCC}
		\toprule
		& \textbf{Type} & \textbf{AD} & \textbf{PI} & \textbf{T1} \\
		\midrule
		Grad CAM \cite{selvaraju2017grad} & Activation & 0.4261 & 0.0749 & 0.4170 \\
		Score CAM \cite{wang2020score} & Activation & 0.2940 & \textbf{0.0915} & 0.5491 \\	
		\midrule
		LRP \cite{bach2015pixel} & Propagation & 0.8333 & 0.2490 & 0.0098 \\
		IG \cite{sundararajan2017axiomatic} & Propagation & 0.8431 & 0.0000 & 1.42e-5 \\
		\midrule
		Occlusion \cite{zeiler2014visualizing} & Perturbation & 0.8429 & 0.0083 & 0.0002 \\
		Ours & Perturbation & \textbf{0.1122} & 0.5843 & \textbf{0.7309} \\
		\bottomrule
	\end{tabularx}
\end{table}

\subsection{Quantitative Comparison}

In quantitative experiments, we selected 120 images from ImageNet-1k dataset to evaluate various interpretability methods by calculating confidence metrics. Before testing, we processed the saliency maps by retaining only the top 40\% of the most salient pixels and masking the rest of input image to observe the resulting change in confidence. Table \ref{tab1} presents the quantitative evaluation results, showing that our proposed method achieved higher confidence scores compared to other baseline methods. Activation-based methods like CAM can effectively highlight decision-critical regions in some samples but may fail in some scenario. Grad-CAM is susceptible to gradient problems, while Score-CAM does not fully account for the interdependencies of feature maps. Propagation-based methods performed poorly, as gradient information is insufficient to fully assess feature importance.

\begin{figure}[!t]
	\centering
	\includegraphics[width=\linewidth]{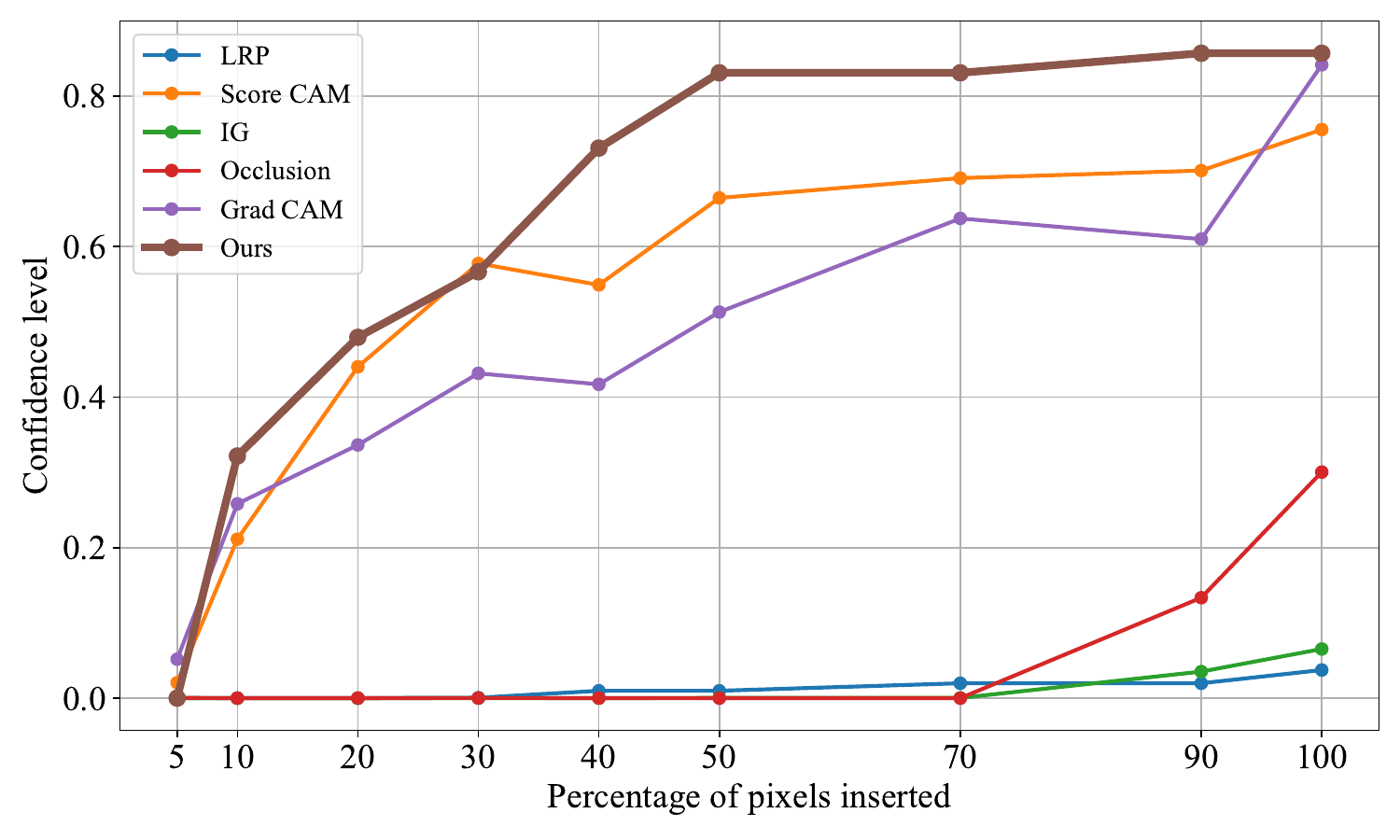}
	\caption{Insertion curves of interpretation methods. Our proposed method only requires minimal pixel information to achieve accurate classification.}
	\label{fig_3}
\end{figure}

We also provide insertion curves to evaluate the effectiveness by retaining a specified percentage of pixels from the saliency maps and calculating the confidence of the masked images. Figure \ref{fig_3} illustrates these insertion curves, showing that our proposed method effectively restores class information even with a small percentage of pixels. When 40\% of the pixels are retained, the method achieves high-accuracy recognition. CAM-based methods can accurately identify targets with minimal information in some samples, but in certain scenarios, incorrect activations make it challenging to pinpoint the regions of network interest.

\subsection{Ablation Study}

\begin{figure}[!t]
	\centering
	\includegraphics[width=\linewidth]{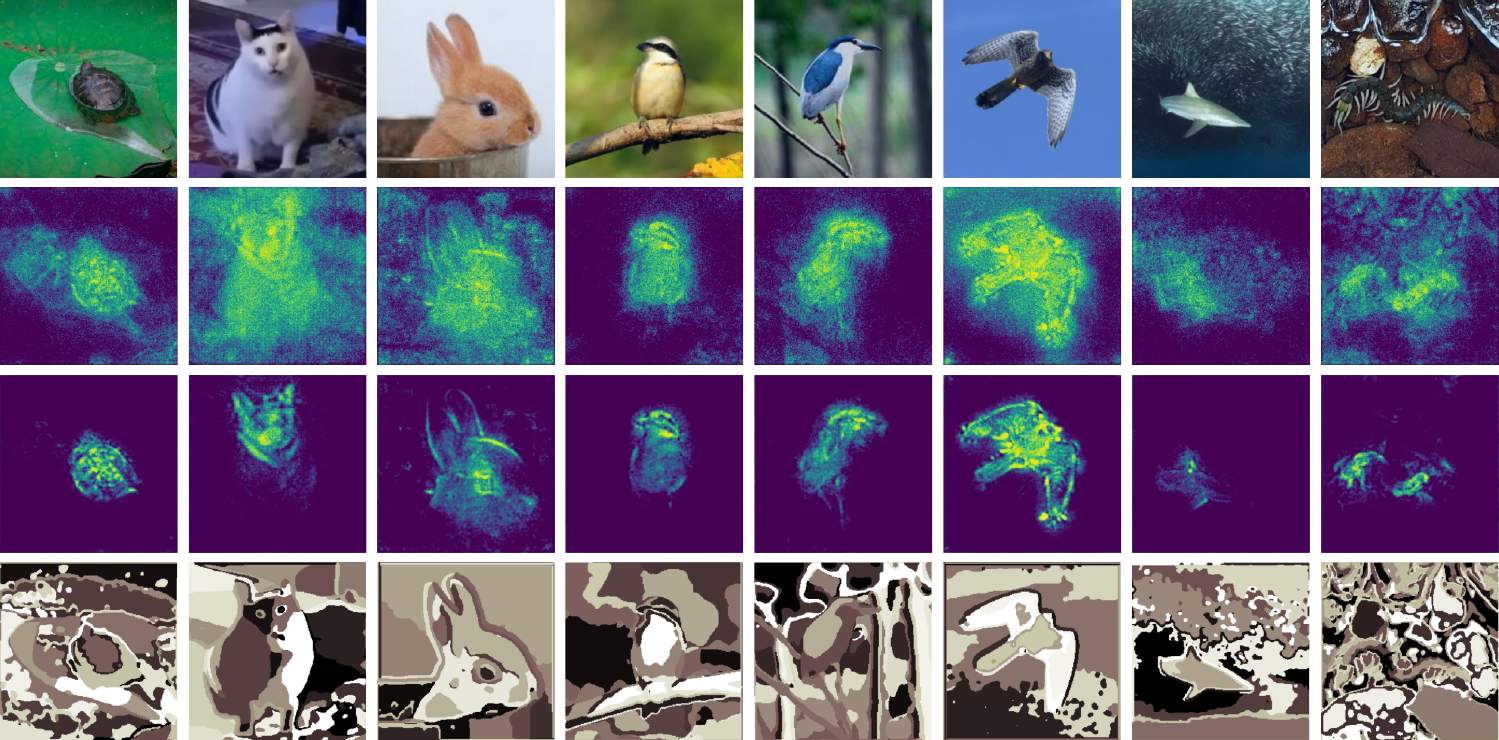}
	\caption{Ablation study. The first row: input samples; The second row: saliency maps without consistency loss; The third row: saliency maps with consistency loss; The fourth row: correlated feature coalitions.}
	\label{fig_4}
\end{figure}

We conducted ablation experiments to validate the effectiveness of loss function designed in our method. Figure \ref{fig_4} shows some visualization results after removing the consistency loss. Resulting saliency maps exhibit more dispersed attention and fail to accurately localize the target. Table \ref{tab2} presents quantitative ablation results for Consistency loss.

\section{Conclusion}

This paper proposed an innovative perturbation-based interpretation that efficiently visualize network decision-making process by extracting relevant feature coalitions from a network perspective. Compared to baseline methods, our approach achieved high confidence level on ImageNet-1k. Additionally, ablation study demonstrated that perturbation on feature coalitions allow interpretation to focus on fine-grained details and edge information that network focus.

\begin{table}[t] 
	\caption{Ablation experiments of Mask loss and Consistency loss.\label{tab2}}
	\scriptsize
	\centering
	\newcolumntype{C}{>{\centering\arraybackslash}X}
	\begin{tabularx}{\linewidth}{CCCC}
		\toprule
		$L_c$ & \textbf{AD} & \textbf{PI} & \textbf{T1} \\
		\midrule
		\texttimes & 0.1887 & 0.0250 & 0.6544 \\
		\checkmark & \textbf{0.1122} & \textbf{0.5843} & \textbf{0.7309} \\
		\bottomrule
	\end{tabularx}
\end{table}

\bibliographystyle{IEEEtran}
\bibliography{IEEEabrv, reference}

\end{document}